\documentclass[11pt]{article}
\usepackage{coling2020}
\usepackage{times}
\usepackage{url}
\usepackage{latexsym}
\usepackage{tcolorbox}
\usepackage{url}
\usepackage{makecell}
\usepackage{amssymb}
\usepackage{amsmath}
\usepackage{graphicx}
\usepackage{array}

\colingfinalcopy % Uncomment this line for the final submission

\title{Two Stage Transformer Model for COVID-19 Fake News Detection and Fact Checking}

\author{
    Rutvik Vijjali\thanks{* These authors contributed equally.} \\
    \small{rutvikvijjali30@gmail.com}\\
    \And
    Prathyush Potluri\footnotemark[1]\\
   \small{potluri.prathyush@gmail.com}\\
    \And 
    Siddharth Kumar\footnotemark[1]\\
   \small{kumar.sidiyer@gmail.com}\\
    \And
    Sundeep Teki\\
   \small{sundeep.teki@gmail.com}
}

\begin{document}

\maketitle
\begin{abstract}
  The rapid advancement of technology in online communication via social media platforms has led to a prolific rise in the spread of misinformation and fake news. Fake news is especially rampant in the current COVID-19 pandemic, leading to people believing in false and potentially harmful claims and stories. Detecting fake news quickly can alleviate the spread of panic, chaos and potential health hazards. We developed a two stage automated pipeline for COVID-19 fake news detection using state of the art machine learning models for natural language processing. The first model leverages a novel fact checking algorithm that retrieves the most relevant facts concerning user claims about particular COVID-19 claims. The second model verifies the level of ``truth” in the claim by computing the textual entailment between the claim and the true facts retrieved from a manually curated COVID-19 dataset. The dataset is based on a publicly available knowledge source consisting of more than 5000 COVID-19 false claims and verified explanations, a subset of which was internally annotated and cross-validated to train and evaluate our models. We evaluate a series of models based on classical text-based features to more contextual Transformer based models and observe that a model pipeline based on BERT and ALBERT for the two stages respectively yields the best results.
\end{abstract}
\section{Introduction}
\blfootnote{
    %
    % for review submission
    %
    % \hspace{-0.65cm}  % space normally used by the marker
    % Place licence statement here for the camera-ready version. See
    % Section~\ref{licence} of the instructions for preparing a
    % manuscript.
    %
    % % final paper: en-uk version 
    %
    % \hspace{-0.65cm}  % space normally used by the marker
    % This work is licensed under a Creative Commons 
    % Attribution 4.0 International Licence.
    % Licence details:
    % \url{http://creativecommons.org/licenses/by/4.0/}.
    % 
    % % final paper: en-us version 
    %
    \hspace{-0.65cm}  % space normally used by the marker
    This work is licensed under a Creative Commons 
    Attribution 4.0 International License.
    License details:
    \url{http://creativecommons.org/licenses/by/4.0/}.
}
{Electronic means of communication have helped to eliminate time and distance barriers to sharing and broadcasting information. However, despite all its advantages, faster means of communication has also resulted in extensive spread of misinformation. The world is currently going through the deadly COVID-19 pandemic and fake news regarding the disease, its cures, its prevention and causes have been broadcast widely to millions of people. The spread of fake news and misinformation during such precarious times can have grave consequences leading to widespread panic and amplification of the threat of the pandemic itself. It is therefore of paramount importance to limit the spread of fake news and ensure that accurate knowledge is disseminated to the public. \newline
In this work, we propose a robust, dynamic fake news detection system, that can not only estimate the “correctness” of a claim but also provides users with pertinent information regarding the said claim. This is achieved using a knowledge base of verified information that can be constantly updated. Previous work on fake news detection has primarily focused on evaluating the relationship measured via a textual entailment task between a header and the body of the article. However, such a method is insufficient for identifying specific fake news claims without any knowledge of the facts relevant to the claim. This warrants the use of a new dataset specific to the COVID-19 pandemic.
%We scraped over 5000 false claims and their relevant explanations from reliable fact checking sources regarding COVID-19 and about 1200 such pairs were manually augmented with the "true claim" as well to train and evaluate the entailment model.
\\
Developing a solution for such a task involves generating a database of factual explanations, which becomes our knowledge base, that serves as ground truth for any given claim. We compute the entailment between any given claim and  explanation to verify if the claim is true or not.
Querying for claim, explanation pairs for each explanation in our knowledge base is computationally expensive and slow, so we propose generating a set of candidate explanations which are contextually similar to the claim. We achieve this by using a model trained with relevant and irrelevant claim explanation pairs, and using a similarity metric between the two to match them. Therefore, in the pipeline, firstly, for a given claim, a set of candidate explanations from the knowledge base have to be fetched in real-time. Then, the claim validated for truth using relevant candidate explanations.  In this work, we have explored the use of Transformer~\cite{vaswani2017attention} based models to both fetch relevant explanations as well as measure the entailment between a given claim and a factual explanation.
\\
We then evaluate our model on the basis of count of relevant explanation results fetched as well as the accuracy in verifying a given claim. We demonstrate the effectiveness of pre-trained multi-attention models in terms of overall accuracy when compared with other natural language processing (NLP) baselines while maintaining near real-time performance.
\blfootnote

}
\section{Related Work}
The paper ~\cite{riedel2017simple} uses traditional approaches with a simple classifier model that makes use of Term Frequency(TF), Term Frequency and Inverse Document Frequency(TF-IDF), and cosine similarity between vectors as features to classify fake news. They have provided a baseline for fake news stance detection on Fake News Challenge (FNC-1) dataset\footnote{http://www.fakenewschallenge.org/}. We have implemented their approach on our dataset and results can be seen in Table~\ref{testresults}.
\\
In ~\cite{nie2019combining}, the authors present a connected system consisting of three homogeneous neural semantic matching models that perform document retrieval, sentence selection, and claim verification on the FEVER Dataset~\cite{thorne2018fever} jointly for fact extraction and verification. Their Neural Semantic Matching Network (NSMN) is a modification of the Enhanced Sequential Inference Model (ESIM)~\cite{chen2017neural}, where they add skip connections from input to matching layer and change output layer to only max-pool plus one affine layer with ReLU activation.
They use three stage pipeline in which given a claim, they first retrieve candidate documents from the corpus, followed by retrieving candidate sentences from the selected candidate documents and finally, the last stage classifies the sentence in to one of three classes. They  used Bidirectional LSTM (BiLSTM) to encode the claim and sentences using GloVe~\cite{pennington2014glove} and ELMO~\cite{Peters:2018} embeddings. However, these tasks are concerned with static or slowly evolving domains, on topics that do not require domain expertise to annotate.\newline
Despite the partial success of the above methods, there were still certain shortcomings in terms of the accuracy of the results. Bidirectional encoder representations from Transformers (BERT)~\cite{devlin2018bert} is a pre-trained language model trained on a large corpus comprising the Wikipedia and Toronto Book Corpus, and is shown to perform well on several natural language tasks like GLUE~\cite{wang2018glue}. Transformer based pretrained models achieved state of the art results in several NLP subtasks, their ease of fine-tuning makes them adaptable to newer tasks. In ~\cite{jwa2019exbake}, the authors propose a model based on the BERT architecture to detect fake news by analyzing the contextual relationship between the headline and the body text of news. They demonstrate that using Transformer based models like BERT and fine-tuning them for the task of fake news detection gives better results than other models like stackLSTM~\cite{hanselowski2018retrospective,hermans2013training} and featMLP~\cite{davis2017fake}. They further enhanced their model performance by pre-training with domain specific news and articles, and countered the class imbalance for the final classification task by the use of a weighted cross entropy loss function. However, this approach can not be adapted for our task of fetching relevant explanations for each claim.\newline
~\cite{naude2020artificial} and ~\cite{bullock2020mapping} extensively studied the use of machine learning strategies to address various issues regarding COVID-19. The latter also describes fake news and misinformation as a major issue in the ongoing pandemic and further highlights the problem as solving an infodemic. In ~\cite{gallotti2020assessing}, the authors develop an Infodemic Risk Index (IRI) after analyzing Twitter posts across various languages and calculate the rate at which a particular user from a locality comes across unreliable posts from different classes of users like verified humans, unverified humans, verified bots, and unverified bots. In ~\cite{mejova2018online}, the authors examine Facebook advertisements across 64 countries and find that around 5\% of advertisements contained possible errors or misinformation. But none of these mentioned works tackle the problem of misinformation by reasoning out the given fake claim with an explanation.
\section{Dataset}
Using an existing misinformation dataset will not serve as a reliable knowledge base for training and evaluating the models due to the recent and uncommon nature i.e., the vocabulary used to describe the disease and the terms associated with the COVID-19 pandemic. It is important to generate real and timely datasets to ensure accurate and consistent evaluation of the methods. To overcome this drawback, we manually curated a dataset specific to COVID-19. Our proposed dataset consists of 5500 claim and explanation pairs. We describe the collection and annotation process in Section~\ref{claimsdataset}.
\subsection{Covid-19 Claims Dataset}
\label{claimsdataset}
There are multiple sources on the web that are regularly identifying and debunking fake news on COVID-19. We scraped data from “Poynter”\footnote{ https://www.poynter.org/ifcn-covid-19-misinformation/} , a fact checking website which collects fake news and debunks or fact-checks them with supporting articles from more than 70 countries, covering more than 40 languages. 
\begin{figure}
\centering
\includegraphics[width=\textwidth]{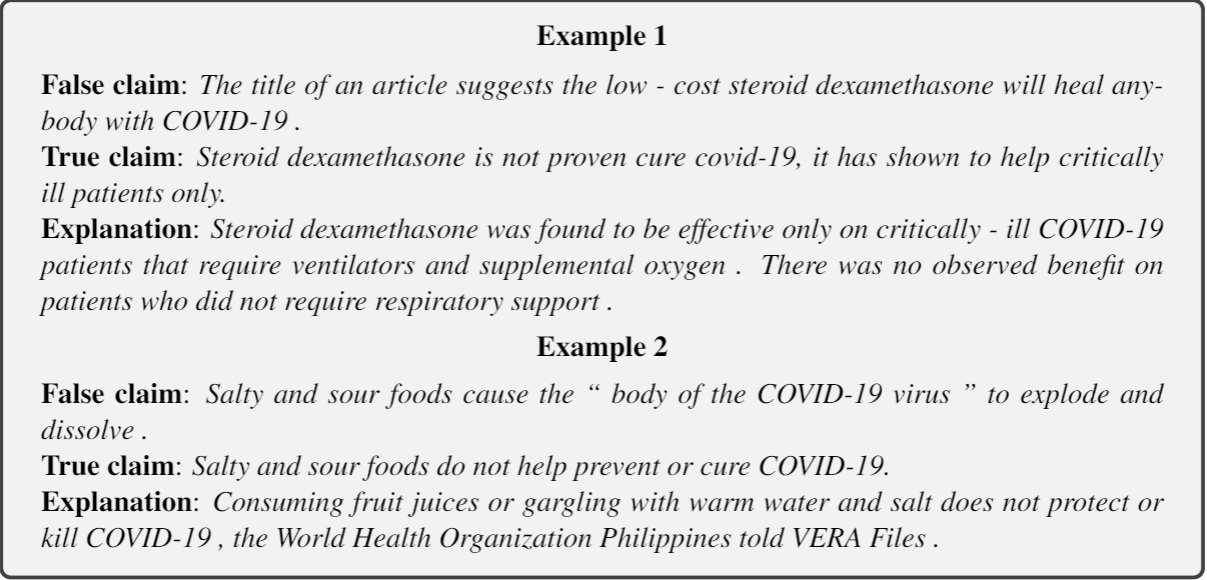}
\caption{Cross validated data examples}
\label{fig:dataexamples}
\end{figure}
The Poynter website has a database exclusively for COVID-19 with over 7000 fact checks. Each fact check contains the corresponding claim that is being checked, the rating indicating the type of the claim, for example - 'False', 'Mostly False' or 'Misleading', the name of the fact checker, the explanation given for the current claim, the location of origin of the claim and the date when the fact check was done. This data can be used to update our "explanation" look-ups in a timely fashion so our database is constantly evolving as we learn more about the virus and the facts change.

% \begin{tcolorbox}
% \begin{center}
%   { \bf{Example 1}}

% \end{center}

% \textbf{False claim}: \textit{The title of an article suggests the low - cost steroid dexamethasone will heal anybody with COVID-19 .} \\
% \textbf{True claim}: \textit{Steroid dexamethasone is not proven cure covid-19, it has shown to help critically ill patients only.}  \\
% \textbf{Explanation}: \textit{Steroid dexamethasone was found to be effective only on critically - ill COVID-19 patients that require ventilators and supplemental oxygen . There was no observed benefit on patients who did not require respiratory support .} 
% \begin{center}
%   { \bf{Example 2}}
% \end{center}
% \textbf{False claim}: \textit{Salty and sour foods cause the “ body of the COVID-19 virus ” to explode and dissolve .}\\
% \textbf{True claim}: \textit{Salty and sour foods do not help prevent or cure COVID-19.} \\
% \textbf{Explanation}: \textit{Consuming fruit juices or gargling with warm water and salt does not protect or kill COVID-19 , the World Health Organization Philippines told VERA Files .}
% \end{tcolorbox}

For each fact check, we collect only the "claim" and the corresponding “explanation” from this database which were rated as 'False' or 'Misleading'. In this way, we collected about 5500 false-claim and explanation pairs. We further manually rephrase these false claims to generate true claims, as the ones that align with the explanation so as to create an equal proportion of true-claim and explanation pairs. We have taken claim-explanation pairs from the time period between January 1,2020 to May 15,2020 for our training data and from May 18,2020 to July 1,2020 for our test data. In this way, we evaluate the generalization of the model on completely new and unseen data. Our collected data follows the structure: 
\begin{center}
   {\it{[false claim, explanation]}} 
\end{center}
The subset of the data that we annotated and cross validated follows the structure: 
\begin{center}
   {\it{[false claim, true claim, explanation]}}
\end{center}
Figure~\ref{fig:dataexamples} shows some examples of the cross validated data.

\subsection{Dataset Statistics}
The current proposed Covid-19 dataset contains cross-validated claim-explanation sentence pairs. Statistics about the distribution of labels are provided in Table~\ref{tab:datastats}. 
This is a dynamic dataset and we are continually collecting and curating additional claim-explanation pairs. We plan to open source this dataset to facilitate more research in this domain.

\begin{table}
\centering
\begin{tabular}{|c|c|}
\hline
\textbf{Dataset}                                                                                                          & \textbf{\begin{tabular}[c]{@{}c@{}}Number of sentence pairs\end{tabular}} \\ \hline
False claim - Explanation pairs                                                                                           & 5500                                                                        \\ \hline
\begin{tabular}[c]{@{}c@{}}Cross validated False claim - True claim - Explanation pairs for train data\end{tabular} & 1000                                                                        \\ \hline
\begin{tabular}[c]{@{}c@{}}Cross validated False claim  - True claim - Explanation pairs for test data\end{tabular}  & 200                                                                         \\ \hline
\end{tabular}
\caption{Dataset Information}
\label{tab:datastats}
\end{table}

\section{Methodology}
\label{method}
The architecture consists of a two stage model, we will refer to the first model as ``Model A" and the second model as ``Model B". The objective of Model A is to fetch the candidate ``true facts" or explanations for a given claim, which are then evaluated for entailment using the Model B. Next, we describe the training procedure as well as intended run time behaviour for both Model A and Model B. 
\begin{figure}[h]
\centering
\includegraphics[width=\textwidth]{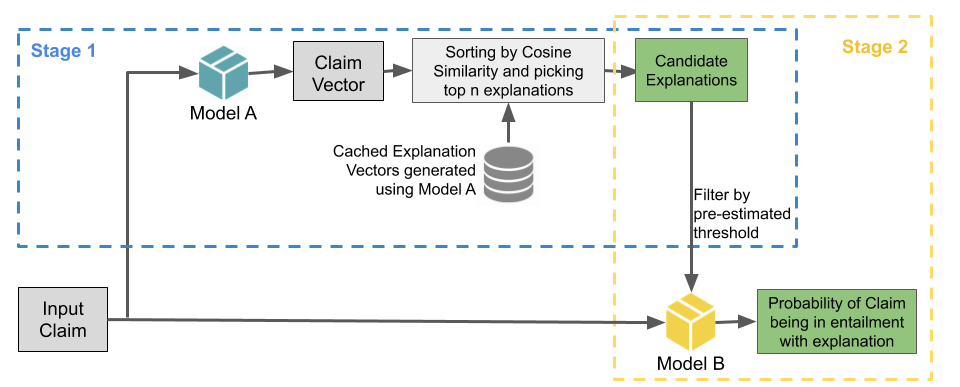}
\caption{Block diagram of our two stage model pipeline}
\label{fig:block_diag}
\end{figure}

\subsection{Model A}
First, to fetch relevant explanations, we train our Transformer model on a binary sentence entailment task, where the claims and explanations are the two sentences fed in as input separated by a [SEP] tag. We generate negative claim-explanation pairs through random sampling to ensure that equal proportions of positive and negative pairs are present. 
Training multi-attention network with our COVID-19 specific data enables the model to capture long-range correlations between the vector representations of claims and explanations of similar contexts. We train our models with a base encoder and a sequence classification head on top for binary classification of the labels. The model is trained to optimise the cross entropy loss. \newline
Through our experiments, we find that, on this trained model, if we generate embeddings for a single sentence (either claim or explanation individually) and compare matching {[claim, explanation]} embeddings using the cosine similarity metric, there is a distinction in the distribution of similarity scores between related and unrelated {[claim,explanation]} pairs. Therefore, for faster near real-time performance, we cache the embeddings for all our explanations (knowledge base) beforehand, and compute the cosine similarity between the claim and the cached embeddings of the explanations. The vector of the [CLS](the start of sentence) token of the final layer works as a strong representation of the entire sentence, although we found that taking element-wise mean over all the token vectors leads to better performance. We fetch the top explanations for any given claim exceeding a certain threshold of sentence similarity as there could be several explanations relevant for a given claim. This threshold is determined on the basis of the summary statistics of the cosine similarity metric between the claim and relevant explanations in the validation set, as described in Section~\ref{eval}. These retrieved explanations serve as candidates for verifying the accuracy of the claim through Model B. 
\subsection{Model B}
The second part of the pipeline is to identify the veracity of a given claim. Model A fetches the candidate explanations while Model B is used to verify whether the given claim aligns with our set of candidate explanations or not. We can therefore treat this task as a textual entailment problem~\cite{dagan2013recognizing,adler2012entailment}. To train the Model B, we use a smaller subset of ``false claim" and ``explanation" pairs from our original dataset, and cross validate each sample with ``true claim" or in other words, claims that align with the factual explanation. However, this small annotated data is not sufficient to train the model effectively. Therefore, the parameters of the Model A, which was trained on a much larger dataset were used as initial parameters for Model B, and fine-tuned further using our cross validated dataset.
\\
We trained Model B in a similar fashion as Model A i.e. as a sequence classification problem with cross entropy loss. Once we have the candidate explanations for a given claim, we use Model B to estimate the probabilities of alignment of claim with each of the candidate explanations. We used the statistic of mean probability score and standard deviation of aligning and non-aligning claim and explanation pairs in the validation set to determine the thresholds for Model B classification. We trained and evaluated both Model A and Model B using several approaches based on classical NLP methods as well as more sophisticated pre-trained Transformer models. The flow of the Model A + Model B pipeline is shown in Figure~\ref{fig:block_diag}.

\section{Experiments}
\subsection{Baseline Models}

For baselining our model on classical NLP approaches, we use two simple bag-of-words (BOW) representations for the text inputs: term frequency (TF) and term frequency-inverse document frequency (TF-IDF). We followed the architecture proposed by ~\cite{riedel2017simple}. The representations and features extracted from the claim and explanation pairs consist of the following:
• The TF vector of the claim;
• The TF vector of the explanation;
• The cosine similarity between the 2-normalised TF-IDF vectors of the claim and explanation.\\
For the TF vectors, we extract a vocabulary of the 5,000 most frequent words in the training set and exclude stop words (the NLTK~\cite{bird2009natural} stop words for the English language). For the TF-IDF vectors, we use the same vocabulary and set of stop words. The TF vectors and the TF-IDF cosine similarity values are concatenated to form a feature vector with total dimension of 10,001 and is fed to the classifier.\newline
The classifier takes an input of 10,001 dimensional vector followed by a feed-forward network with 50, 20 and 2 dimensional dense layers respectively with each hidden layer having tanh activation and the last layer is a softmax layer. We trained in mini-batches over the entire training set using the Adam optimiser~\cite{kingma2014adam} with a learning rate of 0.001.
\\
The second approach involves use of word vectors for which we used 300-dimensional GloVe~\cite{pennington2014glove} embeddings, pretrained on 2014-Wikipedia and Gigaword, and averaged over token embeddings to compute sentence vectors. So, for a given claim and explanation pair, we have a 300-dimension vector for claim as well as the explanation, both of which are concatenated to form a 600-dimensional vector that serves as input to our dense layer classifier. This model is a simple feed-forward neural network with 4 hidden layers having 200, 100, 50 and 2 hidden units respectively with the first three layers having ReLU activation while the last layer is a softmax layer. We trained in mini-batches of 32 over the entire training set with back-propagation using the Adam optimizer with a learning rate of 0.001.

\label{ssec:expts}

\subsection{Transformer Models}
\label{trans}
We trained and evaluated three Transformer based pre-trained models for both Model A and Model B using the training strategy described in Section~\ref{method}. As our focus was to ensure that the proposed pipeline can be deployed effectively in a near real-time scenario, we restricted our experiments to models that could efficiently be deployed using inexpensive compute. We chose the following three models - BERT(base), ALBERT~\cite{lan2019albert} and MobileBERT~\cite{sun2020mobilebert}. The authors of MobileBERT demonstrated that using a teacher-student learning technique for progressive knowledge transfer from the BERT to MobileBERT model helps them achieve a task-agnostic model similar to BERT and can be deployed easily on resource limited devices due to faster inference speeds and lower memory consumption. The ALBERT model was proposed to increase the training and inference speed of BERT besides lowering the memory consumption. The authors demonstrate that the use of their parametric reduction techniques and a custom self supervised loss helps it to achieve results similar to BERT while having fewer parameters. 
\\
Model A was trained on 5000 claim-explanation pairs on the sequence classification task to optimize the softmax cross entropy loss using a learning rate of 3e-5. This trained model was then validated on a test set comprising of 1000 unseen claim-explanation pairs. The training data structure here looks like:
\begin{center}
   {\it{[claim, relevant explanation, 1]}} 
\end{center}
\begin{center}
    {\it{[claim, irrelevant explanation, 0]}}
\end{center}
Model B was trained on a smaller subset of 800 cross validated {[claim,explanation,label]} data, on the same sequence classification task, where the label was assigned based on whether the claim
\begin{table}[]
\centering
\begin{tabular}{|c|c|c|}
\hline
\textbf{Model} & \textbf{\begin{tabular}[c]{@{}c@{}}Model A Val Accuracy\end{tabular}} & \textbf{\begin{tabular}[c]{@{}c@{}}Model B Val Accuracy\end{tabular}} \\ \hline
TF-IDF         & 0.832                                                                   & 0.799                                                                   \\ \hline
GloVe          & 0.781                                                                   & 0.777                                                                   \\ \hline
MobileBERT     & 0.921                                                                   & 0.877                                                                   \\ \hline
ALBERT         & 0.927                                                                   & \textbf{0.956}                                                          \\ \hline
BERT           & \textbf{0.944}                                                          & 0.927                                                                   \\ \hline
\end{tabular}
\caption{Model Performance on Validation Set}
\label{tab:my-table}
\end{table}
aligned with the explanation - 1 or not - 0. This was validated on 200 unseen data-points. The loss function used was softmax cross-entropy with a uniform learning rate of 1e-5. The training data structure here looks like:
\begin{center}
   {\it{[true claim, relevant explanation, 1]}}
\end{center}
\begin{center}
   {\it{[false claim, relevant explanation, 0]}}
\end{center}

\subsection{Evaluation Metrics}
\label{eval}
For evaluating the performance of the overall pipeline model, we first evaluate the performance of Model A in its ability to retrieve relevant explanation. For this we use Mean Reciprocal Rank(MRR)~\cite{craswell2009mean} and Mean Recall @10~\cite{malheiros2012source}, that is the proportion of claims for which the relevant explanation was present in the top 10 most contextual explanation by cosine similarity and their mean inverse rank. Equation~\ref{eqMRR} shows the MRR formula for our evaluation.
\begin{equation}
\label{eqMRR}
\begin{aligned}
MRR = \frac{1}{C}\sum_{i=1}^{C}\frac{1}{rank_{i}}
\end{aligned}
\end{equation}
where $rank_{i}$ is the position of the explanation that is relevant to a particular claim according to test data and $C$ is the total number of claims.\newline
Equation~\ref{eqrec} shows the Recall@10 formula for our evaluation.
\begin{equation}
\label{eqrec}
\begin{aligned}
Recall@10 = \frac{1}{C}\sum_{i=1}^{C}f(\text{top-10 explanations})
\end{aligned}
\end{equation}
\qquad\qquad\qquad\qquad\qquad\qquad\qquad $where $\\
\begin{displaymath}
f(\text{x}) = \begin{cases}
1 &\text{if $true\_exp \in \text{x}$}\\
0 &\text{otherwise}
\end{cases}
\end{displaymath}
Here, $true\_exp$ is the actual relevant explanation for a particular claim according to test data and $C$ is the total number of claims.\newline
Once, Model A has retrieved relevant explanations, we evaluate the performance of Model B on computing the veracity of the claim. Here, we only used explanations that exceed an empirically defined threshold in cosine similarity between the claim and the explanation. Through our experiments, we found that a threshold of \textit{mean - standard deviation of cosine similarity} over the validation data worked well for picking relevant explanations. For evaluating the accuracy, we take a mean of the output probabilities for each $claim,  explanation_{i}$, defined by the Equation~\ref{eq3}. 

\begin{equation}
\label{eq3}
\begin{aligned}
p_{truth}=\frac{1}{n}\sum_{i}modelB(claim,exp_{_{i}}) 
\end{aligned}
\end{equation}
\qquad\qquad\qquad\qquad\qquad\qquad $where $

\qquad\qquad\qquad\qquad\qquad\qquad $exp \in exp_{A} \mid \cos(c\_vec_{A},exp\_vec_{A})>t$\\
\newline
Here, $exp_{A}$ are the top-10 explanations returned by Model A. $c\_vec_{A}$ and $exp\_vec_{A}$ are the vector representations generated by running claim and explanations individually through Model A and $t$ refers to the threshold.
\subsection{Results and Discussion}
The performance of several Transformer based models as well as classical NLP models were compared using the evaluation metrics described in Section~\ref{eval}. The results of the experiments on the test set are summarised in Table~\ref{testresults}.
% TODO: Summarise results, and talk about misclassified claims and invalid explanations fetched with examples.

\begin{table}[h]
\begin{center}
\begin{tabular}{ |p{2.80cm}|p{1.00cm}|p{1.70cm}|p{1.50cm}| } 
\hline
\makecell{\textbf{Model}} & \textbf{MRR} & \textbf{Recall@10} & \textbf{Accuracy}  \\
\hline
TF-IDF  & \makecell{0.477} & \makecell{0.635} & \makecell{{0.525}}\\
\hline
GloVe  & \makecell{0.182} & \makecell{0.410} & \makecell{{0.579}}\\
\hline
MobileBERT  & \makecell{0.561} & \makecell{0.735} & \makecell{0.710}  \\ 
\hline
BERT & \makecell{0.632} & \makecell{0.795} & \makecell{0.810}  \\
\hline
ALBERT & \makecell{0.582} & \makecell{0.675} & \makecell{0.825}\\
\hline
\textbf{BERT+ALBERT} & \makecell{0.632} & \makecell{0.795} & \makecell{0.855}  \\
\hline
\end{tabular}
\caption{Model Performance on test set}
\label{testresults}
\end{center}
\end{table}

\begin{table}[]
\centering
\begin{tabular}{|c|c|c|}
\hline
\textbf{Model}                                          & \textbf{\begin{tabular}[c]{@{}c@{}}Latency \\ per claim\\ (in seconds)\end{tabular}} & \textbf{\begin{tabular}[c]{@{}c@{}}Memory\\ (in MB)\end{tabular}} \\ \hline
TF-IDF                                                  & 0.108                                                                                & 16                                                                \\ \hline
GloVe                                                   & 0.003                                                                                & 990                                                               \\ \hline
MobileBERT                                              & 0.607                                                                                & 1200                                                              \\ \hline
ALBERT                                                  & 2.376                                                                                & 942                                                               \\ \hline
BERT                                                    & 3.106                                                                                & 1910                                                              \\ \hline
\begin{tabular}[c]{@{}c@{}}BERT + ALBERT\end{tabular} & 2.471                                                                                & 1398                                                              \\ \hline
\end{tabular}
\caption{Model Compute performance and Memory usage}
\label{compute}
\end{table}
The results in Table 3 clearly illustrate that Transformer based models are significantly better than classical NLP models. An interesting observation was that some models are better at retrieval of relevant explanations while others have a better classification performance. We find that a combination of the best performing Model A (BERT) and best performing Model B (ALBERT) yielded the highest MRR, Recall@10 and Accuracy on the test set for fact checking. We however do acknowledge that our models could still make errors of two kinds: firstly, Model A might not fetch a relevant explanation which automatically means that the prediction provided by Model B is irrelevant, and secondly, Model A might have fetched the correct explanation(s) but Model B classifies it incorrectly. We show some of the errors our models made in Table~\ref{obser}.\newline
Table~\ref{compute} shows the memory usage and latencies of the implemented models. The memory consumption and latency per claim in the classical NLP models was observed to be quite low in comparison to the Transformer based models. This is expected due to the lower parameter size of the TF-IDF and GloVe models. Among the Transformer based models, MobileBERT had the least latency per claim as expected and explained in Section~\ref{trans} while ALBERT consumed the least memory. The best performing BERT+ALBERT model utilized a memory of 1398MB and fetched relevant explanations of each claim in 2.471 seconds. The model latencies and memory usage were evaluated on an \textit{Intel Xeon - 2.3GHz Single core - 2 thread} CPU.

\begin{table}[htb]
\begin{tabular}{|p{15.40cm}|}
\hline
\begin{tabular}[c]{p{15.00cm}}\textbf{Claim}: \textit{Cannabis could help prevent coronavirus infection . }\\ \textbf{True Explanation}: \textit{The study claiming that marijuana can cure coronavirus did not pass the peer review , it was conducted on artificial human tissues and not on real organisms . It is a classic preliminary research that may even fail . The authors themselves speak of the need for further studies and research .}\\
\textbf{Top Fetched Explanation}: \textit{The vaccine can provide stable immunity , while the presence of antibodies does not prevent reinfection .}\\\textbf{Remark}: \textit{Model A fetched irrelevant explanation for this claim}\\\end{tabular}               \\ \hline
\begin{tabular}[c]{p{15.00cm}}\textbf{Claim}: \textit{The vaccine is not the final solution against the novel coronavirus but antibodies are .}\\ \textbf{Explanation}: \textit{The vaccine can provide stable immunity , while the presence of antibodies does not prevent reinfection .}\\ \textbf{Probability score}: \textit{0.340}\\\textbf{Remark}: \textit{Model B misclassified this claim-explanation pair as True }\\\end{tabular}\\

\hline
\begin{tabular}[c]{p{15.00cm}}\textbf{Claim}: \textit{WHO recommends wearing masks in public spaces to slow down the spread of coronavirus}\\ \textbf{Explanation}: \textit{The WHO changed its position about masks by now recommending community masks in areas with many infections . And it says that masks have to be used properly and alone can't protect you from COVID-19 .}\\
\textbf{Probability score}: \textit{0.686}\\\textbf{Remark}: \textit{Model B misclassified this claim-explanation pair as False}\\\end{tabular}      \\

\hline
\end{tabular}%
\caption{BERT+ALBERT Model Faulty predictions }
\label{obser}
\end{table}

%\textbf{Digital Object Identifiers}:  As part of our work to make ACL
%materials more widely used and cited outside of our discipline, ACL
%has registered as a CrossRef member, as a registrant of Digital Object
%Identifiers (DOIs), the standard for registering permanent URNs for
%referencing scholarly materials.  As of 2017, we are requiring all
%camera-ready references to contain the appropriate DOIs (or as a
%second resort, the hyperlinked ACL Anthology Identifier) to all cited
%works.  Thus, please ensure that you use Bib\TeX\ records that contain
%DOI or URLs for any of the ACL materials that you reference.
%Appropriate records should be found for most materials in the current
%ACL Anthology at \url{http://aclanthology.info/}.

%As examples, we cite \cite{P16-1001} to show you how papers with a DOI
%will appear in the bibliography.  We cite \cite{C14-1001} to show how
%papers without a DOI but with an ACL Anthology Identifier will appear
%in the bibliography.  

\section{Conclusions and Future work}
In this work, we have demonstrated the use and effectiveness of pre-trained Transformer based language models in retrieving and classifying fake news in a highly specialized domain of COVID-19. Our proposed two stage model performs significantly better than other baseline NLP approaches. Our knowledge base, that we prepare through collecting factual data from reliable sources from the web can be dynamic and change to a large extent,  without having to retrain our models again for as long as the distribution is consistent. All of our proposed models can run in near real-time with moderately inexpensive compute. \newline 
Our work is based on the assumption that our knowledge base is accurate and timely. This assumption might not always be true in a scenario such as COVID-19 where ``facts" are changing as we learn more about the virus and its effects. Therefore a more systematic approach is needed for retrieving and classifying claims using this dynamic knowledge base. Our future work consists of weighting our knowledge base on the basis of the duration of the claims and benchmarking each claim against novel sources of ground truth. \newline
Our model performance can be further boosted by better pre-training, through domain specific knowledge. In one of the more recent work by~\cite{guo2020cord19sts}, the authors propose a novel semantic textual similarity dataset specific to COVID-19. Pre-training our models using such specific datasets could help in better understanding of the domain and ultimately better performance. \newline
Fake news and misinformation is an increasingly important and a difficult problem to solve, especially in an unforeseen situation like the COVID-19 pandemic. Leveraging state of the art machine learning and deep learning algorithms along with preparation and curation of novel datasets can help address the challenge of fake news related to COVID-19 and other public health crises.

\bibliographystyle{coling}
\bibliography{coling2020}

\begin{thebibliography}{}

\bibitem[\protect\citename{Adler \bgroup et al.\egroup
  }2012]{adler2012entailment}
Meni Adler, Jonathan Berant, and Ido Dagan.
\newblock 2012.
\newblock Entailment-based text exploration with application to the health-care
  domain.
\newblock In {\em Proceedings of the ACL 2012 System Demonstrations}, pages
  79--84.

\bibitem[\protect\citename{Bird \bgroup et al.\egroup }2009]{bird2009natural}
Steven Bird, Ewan Klein, and Edward Loper.
\newblock 2009.
\newblock {\em Natural language processing with Python: analyzing text with the
  natural language toolkit}.
\newblock " O'Reilly Media, Inc.".

\bibitem[\protect\citename{Bullock \bgroup et al.\egroup
  }2020]{bullock2020mapping}
Joseph Bullock, Katherine~Hoffmann Pham, Cynthia Sin~Nga Lam, Miguel
  Luengo-Oroz, et~al.
\newblock 2020.
\newblock Mapping the landscape of artificial intelligence applications against
  covid-19.
\newblock {\em arXiv preprint arXiv:2003.11336}.

\bibitem[\protect\citename{Chen \bgroup et al.\egroup }2017]{chen2017neural}
Qian Chen, Xiaodan Zhu, Zhen-Hua Ling, Diana Inkpen, and Si~Wei.
\newblock 2017.
\newblock Neural natural language inference models enhanced with external
  knowledge.
\newblock {\em arXiv preprint arXiv:1711.04289}.

\bibitem[\protect\citename{Craswell}2009]{craswell2009mean}
Nick Craswell.
\newblock 2009.
\newblock Mean reciprocal rank.
\newblock {\em Encyclopedia of database systems}, 1703.

\bibitem[\protect\citename{Dagan \bgroup et al.\egroup
  }2013]{dagan2013recognizing}
Ido Dagan, Dan Roth, Mark Sammons, and Fabio~Massimo Zanzotto.
\newblock 2013.
\newblock Recognizing textual entailment: Models and applications.
\newblock {\em Synthesis Lectures on Human Language Technologies}, 6(4):1--220.

\bibitem[\protect\citename{Davis and Proctor}2017]{davis2017fake}
Richard Davis and Chris Proctor.
\newblock 2017.
\newblock Fake news, real consequences: Recruiting neural networks for the
  fight against fake news.

\bibitem[\protect\citename{Devlin \bgroup et al.\egroup }2018]{devlin2018bert}
Jacob Devlin, Ming-Wei Chang, Kenton Lee, and Kristina Toutanova.
\newblock 2018.
\newblock Bert: Pre-training of deep bidirectional transformers for language
  understanding.
\newblock {\em arXiv preprint arXiv:1810.04805}.

\bibitem[\protect\citename{Gallotti \bgroup et al.\egroup
  }2020]{gallotti2020assessing}
Riccardo Gallotti, Francesco Valle, Nicola Castaldo, Pierluigi Sacco, and
  Manlio De~Domenico.
\newblock 2020.
\newblock Assessing the risks of" infodemics" in response to covid-19
  epidemics.
\newblock {\em arXiv preprint arXiv:2004.03997}.

\bibitem[\protect\citename{Guo \bgroup et al.\egroup }2020]{guo2020cord19sts}
Xiao Guo, Hengameh Mirzaalian, Ekraam Sabir, Aysush Jaiswal, and Wael
  Abd-Almageed.
\newblock 2020.
\newblock Cord19sts: Covid-19 semantic textual similarity dataset.
\newblock {\em arXiv preprint arXiv:2007.02461}.

\bibitem[\protect\citename{Hanselowski \bgroup et al.\egroup
  }2018]{hanselowski2018retrospective}
Andreas Hanselowski, Avinesh PVS, Benjamin Schiller, Felix Caspelherr, Debanjan
  Chaudhuri, Christian~M Meyer, and Iryna Gurevych.
\newblock 2018.
\newblock A retrospective analysis of the fake news challenge stance detection
  task.
\newblock {\em arXiv preprint arXiv:1806.05180}.

\bibitem[\protect\citename{Hermans and Schrauwen}2013]{hermans2013training}
Michiel Hermans and Benjamin Schrauwen.
\newblock 2013.
\newblock Training and analysing deep recurrent neural networks.
\newblock In {\em Advances in neural information processing systems}, pages
  190--198.

\bibitem[\protect\citename{Jwa \bgroup et al.\egroup }2019]{jwa2019exbake}
Heejung Jwa, Dongsuk Oh, Kinam Park, Jang~Mook Kang, and Heuiseok Lim.
\newblock 2019.
\newblock exbake: Automatic fake news detection model based on bidirectional
  encoder representations from transformers (bert).
\newblock {\em Applied Sciences}, 9(19):4062.

\bibitem[\protect\citename{Kingma and Ba}2014]{kingma2014adam}
Diederik~P Kingma and Jimmy Ba.
\newblock 2014.
\newblock Adam: A method for stochastic optimization.
\newblock {\em arXiv preprint arXiv:1412.6980}.

\bibitem[\protect\citename{Lan \bgroup et al.\egroup }2019]{lan2019albert}
Zhenzhong Lan, Mingda Chen, Sebastian Goodman, Kevin Gimpel, Piyush Sharma, and
  Radu Soricut.
\newblock 2019.
\newblock Albert: A lite bert for self-supervised learning of language
  representations.
\newblock {\em arXiv preprint arXiv:1909.11942}.

\bibitem[\protect\citename{Malheiros \bgroup et al.\egroup
  }2012]{malheiros2012source}
Yuri Malheiros, Alan Moraes, Cleyton Trindade, and Silvio Meira.
\newblock 2012.
\newblock A source code recommender system to support newcomers.
\newblock In {\em 2012 IEEE 36th Annual Computer Software and Applications
  Conference}, pages 19--24. IEEE.

\bibitem[\protect\citename{Mejova \bgroup et al.\egroup
  }2018]{mejova2018online}
Yelena Mejova, Ingmar Weber, and Luis Fernandez-Luque.
\newblock 2018.
\newblock Online health monitoring using facebook advertisement audience
  estimates in the united states: evaluation study.
\newblock {\em JMIR public health and surveillance}, 4(1):e30.

\bibitem[\protect\citename{Naud{\'e}}2020]{naude2020artificial}
Wim Naud{\'e}.
\newblock 2020.
\newblock Artificial intelligence against covid-19: An early review.

\bibitem[\protect\citename{Nie \bgroup et al.\egroup }2019]{nie2019combining}
Yixin Nie, Haonan Chen, and Mohit Bansal.
\newblock 2019.
\newblock Combining fact extraction and verification with neural semantic
  matching networks.
\newblock In {\em Proceedings of the AAAI Conference on Artificial
  Intelligence}, volume~33, pages 6859--6866.

\bibitem[\protect\citename{Pennington \bgroup et al.\egroup
  }2014]{pennington2014glove}
Jeffrey Pennington, Richard Socher, and Christopher~D Manning.
\newblock 2014.
\newblock Glove: Global vectors for word representation.
\newblock In {\em Proceedings of the 2014 conference on empirical methods in
  natural language processing (EMNLP)}, pages 1532--1543.

\bibitem[\protect\citename{Peters \bgroup et al.\egroup }2018]{Peters:2018}
Matthew~E. Peters, Mark Neumann, Mohit Iyyer, Matt Gardner, Christopher Clark,
  Kenton Lee, and Luke Zettlemoyer.
\newblock 2018.
\newblock Deep contextualized word representations.
\newblock In {\em Proc. of NAACL}.

\bibitem[\protect\citename{Riedel \bgroup et al.\egroup
  }2017]{riedel2017simple}
Benjamin Riedel, Isabelle Augenstein, Georgios~P Spithourakis, and Sebastian
  Riedel.
\newblock 2017.
\newblock A simple but tough-to-beat baseline for the fake news challenge
  stance detection task.
\newblock {\em arXiv preprint arXiv:1707.03264}.

\bibitem[\protect\citename{Sun \bgroup et al.\egroup }2020]{sun2020mobilebert}
Zhiqing Sun, Hongkun Yu, Xiaodan Song, Renjie Liu, Yiming Yang, and Denny Zhou.
\newblock 2020.
\newblock Mobilebert: a compact task-agnostic bert for resource-limited
  devices.
\newblock {\em arXiv preprint arXiv:2004.02984}.

\bibitem[\protect\citename{Thorne \bgroup et al.\egroup }2018]{thorne2018fever}
James Thorne, Andreas Vlachos, Christos Christodoulopoulos, and Arpit Mittal.
\newblock 2018.
\newblock Fever: a large-scale dataset for fact extraction and verification.
\newblock {\em arXiv preprint arXiv:1803.05355}.

\bibitem[\protect\citename{Vaswani \bgroup et al.\egroup
  }2017]{vaswani2017attention}
Ashish Vaswani, Noam Shazeer, Niki Parmar, Jakob Uszkoreit, Llion Jones,
  Aidan~N Gomez, {\L}ukasz Kaiser, and Illia Polosukhin.
\newblock 2017.
\newblock Attention is all you need.
\newblock In {\em Advances in neural information processing systems}, pages
  5998--6008.

\bibitem[\protect\citename{Wang \bgroup et al.\egroup }2018]{wang2018glue}
Alex Wang, Amanpreet Singh, Julian Michael, Felix Hill, Omer Levy, and Samuel~R
  Bowman.
\newblock 2018.
\newblock Glue: A multi-task benchmark and analysis platform for natural
  language understanding.
\newblock {\em arXiv preprint arXiv:1804.07461}.

\end{thebibliography}

\end{document}